\documentclass[twoside,leqno,twocolumn]{article}
\usepackage{url}

\usepackage{graphicx,amssymb,amsmath,amsthm}
\usepackage{algorithm}
\usepackage[noend]{algpseudocode}
\usepackage[margin=2.5cm]{geometry}
\usepackage[UKenglish]{isodate}
\usepackage{makecell}
\usepackage{authblk}

\usepackage{subcaption}
\usepackage[font=small,labelfont=bf, justification=justified, format=plain]{caption}
\newcommand{\indi}[1]{\mathbf{1} \lbrace #1 \rbrace}
\newcommand{\D}{\mathsf{D}}
\newcommand{\bA}{{\bf A}}

\newcommand{\bI}{{\bf I}}

\newcommand{\bx}{{\boldsymbol x}}

\newcommand{\bX}{{\boldsymbol X}}

\newcommand{\bw}{{\boldsymbol w}}

\newcommand{\NN}{\mathrm{NN}}
\newcommand{\maj}{\mathrm{maj}}

\DeclareMathOperator{\clust}{c}

\makeatletter
\def\BState{\State\hskip-\ALG@thistlm}
\makeatother

\begin{document}

\title{Nearest Neighbor Median Shift Clustering for Binary Data}
\date{}

\author[1]{Ga\"el Beck}
\author[1]{Tarn Duong }
\author[1]{Mustapha Lebbah }
\author[1]{Hanane Azzag }

\affil[1]{Computer Science Laboratory of Paris North (LIPN, CNRS UMR 7030), University of Paris 13, F-93430 Villetaneuse France}

\cleanlookdateon
\maketitle

\begin{abstract}
We describe in this paper the theory and practice behind a new modal clustering method for binary data. Our approach (BinNNMS) is based on the nearest neighbor median shift. The median shift is an extension of the well-known mean shift, which was designed for continuous data, to handle binary data. We demonstrate that BinNNMS can discover accurately the location of clusters in binary data with theoretical and experimental analyses.

\end{abstract}

\section{Introduction}

The goal of clustering (unsupervised learning) is to assign cluster membership to unlabeled candidate points where the number and location of these clusters are unknown. 
Clustering is an important step in the exploratory phase of data analysis, and it becomes more difficult when applied to binary or mixed data.  Binary data occupy a special place in 
many application
fields: behavioral and social research, survey analysis,
document clustering, and inference on binary images.
%

Clusters are formed usually from a process that minimizes the dissimilarities inside the clusters and to maximizes the dissimilarities between clusters.   
A popular clustering algorithm for binary data is the $k$-modes \cite{Huang97clusteringlarge}, and it is similar to the $k$-means clustering \cite{macqueen1967} wherein the modes are used instead of the means for the prototypes of the clusters. 
Other clustering algorithms have been developed 
using a matching dissimilarity measure for categorical
points instead of Euclidean distance \cite{Li2006}, and a frequency-based method to update
modes in the clustering process \cite{LebbahBT00}.

 %
In this paper, we focus on the mean shift clustering \cite{fukunaga1973,fukunaga1975}, which is another generalization of the $k$-means clustering. Mean shift clustering belongs to the class of modal clustering methods where the arbitrarily shaped clusters are defined in terms of the 
basins of attraction to the local modes of the data density, created by the density gradient ascent paths. In the traditional characterization of the mean shift, these gradient ascent paths are computed from successive iterations of the mean of the nearest neighbors of the current prototype.  Due to its reliance on mean computations, it is not suited to be directly applied to binary data. 
Our contribution is the presentation of a modified mean shift clustering which is adapted to binary data. It is titled Nearest Neighbor Median Shift clustering for binary data (BinNNMS). The main novelty is the that the cluster prototypes are updated via iterations on the majority vote of their nearest neighbors.
We demonstrate that this majority vote corresponds to the median of the nearest neighbors with respect to the Hamming distance \cite{Hamming50}. 
Implementation of the algorithm is available in scala at \url{https://github.com/Clustering4Ever/Clustering4Ever}

The rest of the paper is organized as follows:
Section~\ref{sec:mean-shift} introduces the traditional mean-shift algorithm for continuous data, Section~\ref{sec:nnmsbin} presents our new median shift clustering procedure for binary data BinNNMS, and Section~\ref{sec:exp} 
describes the results of the BinNNMS compared to the $k$-modes clustering. 



\section{Nearest neighbor mean shift clustering for continuous data}
\label{sec:mean-shift}
The mean shift clustering proceeds in an indirect manner based on local gradients of the data density, and
without imposing an ellipsoidal shape to clusters or that the number of clusters be known, as is the case for $k$-means clustering. 
For a candidate point $\bx$, 
the theoretical mean shift recurrence relation is
\begin{equation}
\bx_{j+1} = \bx_j + \frac{\bA \D f(\bx_j)}{f(\bx_j)}
\label{eq:ms}
\end{equation}
for a given positive-definite matrix $\bA$, for $j \geq 1$  and $\bx_0=\bx$.
The output from Equation~\eqref{eq:ms} is the sequence $\{ \bx_j \}_{j\geq0}$ which follows the density gradient ascent $\D f$ to a local mode of the density function $f$. 

To derive the formula for the nearest neighbor mean shift for a random sample $\bX_1, \dots, \bX_n$ drawn from a common density $f$, we replace the 
density $f$ and density gradient $\D f$ by their nearest neighbor estimates
$$
\begin{aligned}
\hat{f}_\NN(\bx; k) &= n^{-1} \delta_{(k)}(\bx)^{-d}
\sum_{i=1}^n \frac{K((\bx - \bX_i)}{\delta_{(k)}(\bx))} \nonumber \\
\D \hat{f}_\NN (\bx; k) 
&= n^{-1}  \delta_{(k)}(\bx)^{-d-1}
\sum_{i=1}^n \frac{\D K((\bx - \bX_i)}{\delta_{(k)}(\bx))}
\end{aligned}
$$
where $K$ is a kernel function and $\delta_{(k)}(\bx)$
as the $k$-th nearest neighbor distance to $\bx$, i.e.
$\delta_{(k)}(\bx)$ is the $k$-th order statistic of the Euclidean 
distances $\lVert \bx- \bX_1 \Vert, \dots, \lVert \bx- \bX_n \Vert$. 
These nearest neighbor estimators were 
introduced by \cite{loftsgaarden1965} and elaborated by 
\cite{fukunaga1973,fukunaga1975} for the mean shift. 

These authors established that the
beta family kernels are computationally efficient for estimating 
$f$ and $\D f$ for continuous data. The uniform kernel is the most widely known member of this beta family, and it is defined as $K(\bx) =  v_0^{-1} \indi{\bx \in B_d(\pmb 0, 1)}$ where $B_d (\bx, r)$ is the $d$-dimensional hyper-ball centered at $\bx$ with radius $r$ and $v_0$ is the hyper-volume of the unit $d$-dimensional hyper-ball $B_d(\pmb 0, 1)$. 
With this family of kernels, and the choice $\bA = (d+2)^{-1} \delta_{(k)}(\bx) \bI_d$, the nearest neighbor mean shift becomes
\begin{align}
\label{eq:nnms}
\bx_{j+1} = k^{-1} \sum_{\bX_i \in \NN_k(\bx_j)} \bX_i
\end{align}
where $\NN_k(\bx)$ is the set of the $k$ nearest neighbors of $\bx$. For the derivation
of Equation~\eqref{eq:nnms}, see \cite{fukunaga1975,duong2016prl}. This nearest neighbor mean shift has a simple interpretation since in the mean shift recurrence relation,
the next iterate $\bx_{j+1}$ is the sample mean of  
the $k$ nearest neighbors of the current iterate $\bx_j$. On the other hand, as these iterations calculate the sample mean, 
the mean shift is not directly applicable to binary data.

\section{Nearest neighbor median shift clustering for binary data}
\label{sec:nnmsbin}
A categorical feature, which has a finite (usually small) number of possible values, can be represented by a binary vector, i.e. a vector which is composed solely of zeroes and ones. These categorical features can either ordinal (which have an implicit order) or can be nominal (which no order exists). Table~\ref{table:modalite} presents the two main types of the coding for a categorical feature into a binary vector, additive and disjunctive, for an example of 3-class categorical feature.

\begin{table}[htbp]
\centering
\begin{tabular}{|c|c|c|}
\hline
Class & {\small Additive coding} & {\small Disjunctive coding}\\
\hline
 {\small 1}&{\small  1 0 0} &{\small  1 0 0}\\
2 & {\small 1 1 0} &{\small  0 1 0} \\
 3 & {\small 1 1 1} &{\small  0 0 1}\\
\hline
\end{tabular}
\caption {Additive and disjunctive coding for a 3-class categorical feature.}
\label{table:modalite}
\end{table} 


The usual Euclidean distance is not adapted to measuring the dissimilarities between binary vectors. A popular alternative is the Hamming distance  $\cal H $ \cite{Leich98}. The Hamming distance between two binary 
vectors $\bx_1=(x_{11}, \dots,x_{1d})$  and  $\bx_2=(x_{21}, \dots,x_{2d})$, $\bx_j \in \{0,1\}^d, j \in 1,2,$ is defined as: 
\begin{align}
{\cal H}(\bx_1,\bx_2)
&=\sum_{j=1}^{d} |x_{1j} -x_{2j}|\nonumber \\
&= d-(\bx_1 - \bx_2)^\top (\bx_1 - \bx_2).
\label{eqn:h}
\end{align}
Equation~\eqref{eqn:h} measures the number of mismatches between the two vectors $\bx_1$ and $\bx_2$:
as the inner product $(\bx_1 - \bx_2)^\top (\bx_1 - \bx_2)$ counts the number of elements
which agree in both $\bx_1$ and $\bx_2$, then $d-(\bx_1 - \bx_2)^\top (\bx_1 - \bx_2)$ counts the number of disagreements. 

The Hamming distance is the basis from which we define the median center of a set of
observations ${\mathcal X} = \{\bX_1, \dots, \bX_n\}, \bX_i \in \{0,1\}^d, i=1,\dots, n$. 
Importantly the median center of the set of binary vectors, as a measure of the centrality of the values, remains a binary vector, unlike the mean vector which can take on intermediate values. 
The median center of ${\mathcal X}$ is a point $\bw=(w_1, \dots,w_d)$
which minimizes the inertia of ${\mathcal X}$, i.e. 
\begin{align}
\label{eqn:cost}
\bw &= \underset{\bx \in \{0,1\}^d}{\mathrm{argmin}} \ \mathcal{I}(\bx)
\end{align}
where
$\displaystyle \mathcal I (\bx) =  \sum_{i=1}^n \pi_i {\cal H}(\bX_i,\bx) = \sum_{i=1}^n \sum_{j=1}^d  \pi_i \mathcal I (x_j)$,  
$\pi_i$ are the weights and $\mathcal I (x_j) = |X_{ij} - x_j|$.
Each component $w_j$ of $\bw$ minimizes
$\mathcal I (x_j)$. 

In the case where all the weights are set to 1, $\pi_i=1, i=1,\dots,n $,
the $w_j$ can be easily computed since it is the most common value in the observations of the $j$-th feature. This is denoted
as $\maj(\mathcal X)$, the component-wise majority vote winner among the data points.
Hence the median center is the majority vote, $\bw = \maj(\mathcal{X})$. 

If we minimize the cost function in Equation~\eqref{eqn:cost} using the dynamic clusters \cite{Diday1976} then this 
leads to the $k$-modes clustering.
Like the $k$-means algorithm, the $k$-modes operates in two steps: (a) an assignment step which assigns each candidate point $\bx$ to the nearest cluster with respect to the Hamming distance, and (b) an optimization
step which computes the median center as the majority vote. These two steps are executed iteratively until the
value of $\mathcal{I}(\bx)$ converges.

Now we show how the median center can be utilized to define a new modal clustering for binary data based on the mean shift paradigm.  In Section~\ref{sec:mean-shift}, the beta family kernels were used in the mean shift for
continuous data. 
The most commonly used smoothing kernel, introduced by \cite{aitchinson1976}, for binary data 
is the Aitchison and Aitken kernel: 
$$K_{\lambda}(\bx) = \lambda^{d- \bx^\top \bx} (1-\lambda)^{\bx^\top \bx}, \ \bx \in \{0,1\}^d.$$ 
Observe that the exponent for $\lambda$ is the Hamming distance of $\bx$. 
The tuning parameter $\tfrac12 \leq \lambda \leq 1$ controls the spread of the probability mass around the origin $\pmb 0$. 
For $\lambda=1/2$, then 
$K_{1/2} (\bx) = (1/2)^d$,  
which assigns a constant probability to 
all points $\bx$, regardless of its distance from $\pmb 0$.  
For $\lambda=1$, $K_1(\bx) = \indi{\bx = \pmb 0}$, 
which assigns all the probability mass to $\pmb 0$. 
For intermediate values of $\lambda$, we have intermediate assignment of between point and uniform probability mass.  

Using $K_\lambda$, the corresponding 
kernel density estimate is 
\begin{align}
\tilde{f}(\bx; \lambda) 
&= n^{-1} \sum_{i=1}^n \lambda^{[d - (\bx - \bX_i)^\top (\bx - \bX_i)]} \nonumber\\
&\quad \cdot
(1-\lambda)^{[(\bx - \bX_i)^\top (\bx - \bX_i)]}.
\label{eq:ftilde}
\end{align}
Since the gradient of the kernel $K_\lambda$ is 
$\D K_\lambda(\bx) 
= 2\bx  \log ((1-\lambda)/\lambda) K_\lambda(\bx)$, 
the density gradient estimate is 
\begin{align}
&\D \tilde{f}(\bx; \lambda) 
=  2 \log (\lambda/(1-\lambda)) n^{-1} \nonumber  \\
&\quad \cdot  \bigg[\sum_{i=1}^n \bX_i  K_\lambda(\bx - \bX_i)
- \bx \sum_{i=1}^n K_\lambda(\bx - \bX_i)\bigg]. 
\label{eq:Dftilde}
\end{align}

To progress in our development of a nearest neighbor median shift for binary data, 
we focus on the point mass kernel $K_1 (\bx) = \indi{\bx = \pmb 0}$.  
In order that ensure that it is amenable for the median shift, we 
modify $K_1$ with two main changes:
\begin{enumerate}
 \item $K_1$ is multiplied by the indicator function $\indi{\bx \in B_d(\pmb 0, 1)}$
\item the indicator function $\indi{\bx = \pmb 0}$, which places the point mass at the center $\pmb 0$, is replaced an indicator that places it on $\maj(B_d(\pmb 0, 1))$, where $\maj(B_d(\pmb 0, 1))$ is the majority vote winner/median center of the data points $\bX_1, \dots, \bX_n$ inside of $B_d(\pmb 0, 1)$.
\end{enumerate}
This second modification results in an asymmetric kernel  as the point mass is no longer always placed in the
centre of the unit ball. 
This modified, asymmetric kernel $L$ is  
$$L(\bx) 
= \indi{\bx = \maj(B_d(\pmb 0, 1))} 
\indi{\bx \in B_d(\pmb 0, 1)}. 
$$

Since $L$ is not directly differentiable, we define its derivative indirectly via $\D K_1$ and  
the convention that $\log (\lambda/(1-\lambda)) = 1$ for $\lambda=1$. 
As $ \D K_\lambda(\bx) \big\lvert_{\lambda=1} = 2 \bx K_1 (\bx)$ then analogously we define  
$\D L(\bx) = 2 \bx L(\bx)$.
To obtain the corresponding estimators, we substitute $L, \D L$ for $K, \D K$ in $\tilde{f}, \D \tilde{f}$ in Equations~\eqref{eq:ftilde}--\eqref{eq:Dftilde} to obtain $\hat{f}, \D \hat{f}$:
\begin{align}
\begin{split}
\hat{f}(\bx;k) &= n^{-1} \delta_{(k)}(\bx)^{-d} \sum_{i=1}^n L((\bx - \bX_i)/\delta_{(k)}(\bx)) \nonumber \\
\D \hat{f}(\bx;k) 
&= 2 \delta_{(k)}(\bx)^{-d-1} n^{-1} \\
&\quad \cdot \bigg[ \sum_{i=1}^n \bX_i  L((\bx - \bX_i)/\delta_{(k)}(\bx))\\
&\quad - \bx  \sum_{i=1}^n L((\bx - \bX_i)/\delta_{(k)}(\bx)) \bigg].
\end{split}
\end{align}

To obtain a nearest neighbor mean shift recurrence relation for binary data, we substitute $\hat{f}, \D \hat{f}$ for $f, \D f$ is Equation~\eqref{eq:ms}.  
For these estimators, the appropriate choice of $\bA = \tfrac12 \delta_{(k)}(\bx) \bI_d$.
Then we have
\begin{align*}
\bx_{j+1} &= \bx_j + \frac{\delta_{(k)}(\bx)}{2} \frac{\D \hat{f}(\bx_j;k)}{\hat{f}(\bx_j; k)} \\ 
&= \frac{ \sum_{i=1}^n \bX_i  L((\bx_j - \bX_i)/ \delta_{(k)}(\bx_j))}{\sum_{i=1}^n L((\bx_j - \bX_i)/ \delta_{(k)}(\bx_j))}.
\end{align*}
We can simplify this ratio if we observe that the scaled kernel is 
\begin{align*}
&L((\bx - \bX_i)/\delta_{(k)}(\bx)) =
\indi{\bX_i \maj(B_d(\bx, \delta_{(k)}(\bx)))} \\
&\quad \cdot \indi{\bX_i \in B_d(\bx, \delta_{(k)}(\bx))};
\end{align*}
and that $B_d(\bx, \delta_{(k)}(\bx))$ comprises the $k$ nearest neighbors of $\bx$, 
then $\indi{\bX_i \in B_d(\bx, \delta_{(k)}(\bx))} = \indi{\bX_i \in \NN_k(\bx)}$. If $m$ is the number of nearest neighbors of $\bx_j$ which coincide with the majority vote, then 
\begin{align}
\label{eq:nnms-bin}
\bx_{j+1} &= \frac{\sum_{\bX_i \in \NN_k(\bx_j)} \bX_i \indi{\bX_i = \maj(\NN_k(\bx_j))}}{\sum_{\bX_i \in \NN_k(\bx_j)} \indi{\bX_i = \maj(\NN_k(\bx_j))}} \nonumber\\ 
&= \frac{m\cdot
\maj(\NN_k(\bx_j))}{m} \nonumber\\
&=  \maj(\NN_k(\bx_j)).
\end{align}
Therefore in the median shift recurrence relation in Equation~\eqref{eq:nnms-bin},
the next iterate $\bx_{j+1}$ is the median center of 
the $k$ nearest neighbors of the current iterate $\bx_j$.    
Thus, once the binary gradient ascent has terminated, the converged point can be decoded using Table~\ref{table:modalite}), allowing 
for its unambiguous symbolic interpretation. 
%
The gradient ascent paths towards the local modes produced by Equation~\eqref{eq:nnms-bin}
form the basis of Algorithm~\ref{alg:nnga}, our  
nearest neighbor median shift clustering for binary data method (BinNNMS).

The inputs to BinNNMS are the data sample $\bX_1, \dots, \bX_n$ and the candidate points 
$\bx_1, \dots, \bx_m$ which we wish to cluster (these can be the same
as $\bX_1, \dots, \bX_n$, but this is not required); and the tuning parameters: 
the number of nearest neighbors $k_1$ used in BGA task, 
the maximum number of iterations $j_{\max}$, and the tolerance under which two cluster centres are considered form a single cluster $\varepsilon$. The output are the cluster labels of the candidate points $\lbrace \clust(\bx_1), \dots, \clust(\bx_m) \rbrace$. 

The aim of the $\varepsilon$-proximity cluster labeling step is to gather all points which are under a threshold  $\varepsilon$. 
In order to apply this method we have to build the Hamming similarity matrix which has a $O(n^2)$ time complexity. 
We initialize the process by taking first point and cluster with it all point whose distance is less than $\varepsilon$. Thus we apply this iterative exploration process 
by adding the  nearest neighbors. 
Once the first cluster is generated, we take another point from the reduced similarity matrix and repeat the process, until all points are assigned a cluster label. A notable problem still remains with the choice of main tuning parameter $\varepsilon$: 
we set it to be the average of distance from each point to their $k_2$ nearest neighbors.

\begin{algorithm}[H]
\caption{BinNNMS -- Nearest neighbor median shift clustering for binary data}
\label{alg:nnga}
\begin{algorithmic}[1]
\Statex {\bf Input:} $\lbrace \bX_1, \dots, \bX_n \rbrace, \lbrace \bx_1, \dots, \bx_m \rbrace, k_1,k_2$, $j_{\max}$ 
\Statex {\bf Output:} $\lbrace \clust(\bx_1), \dots, \clust(\bx_m)\rbrace$
\Statex /* {\bf BGA task}: compute binary gradient ascent paths */
\For{$\ell := 1$ to $m$}
\State $j := 0$; $\bx_{\ell,0} :=\bx_\ell$; 
\State $\bx_{\ell,1} := $ $\maj(\NN_{k_1}(\bx_{\ell,0}))$;
\While{ $j < j_{\mathrm{max}}$}
\State $j := j+ 1$; 
\State $\bx_{\ell,j+1} := $ $\maj(\NN_{k_1}(\bx_{\ell,j}))$;
\EndWhile
\State{$\bx_\ell^* := \bx_{\ell,j}$;}
\EndFor
\Statex /* {\bf $\varepsilon$-proximity cluster labeling task}:  create clusters by merging near final iterates*/
\For{$\ell_1, \ell_2 := 1$ to $m$}
\If{${\cal H}(\bx_{\ell_1}^*,\bx_{\ell_2}^*) \, \leq \varepsilon(k_2)$}
{$\clust(\bx_{\ell_1}^*) := \clust(\bx_{\ell_2}^*)$;}
\EndIf
\EndFor
\end{algorithmic}
\end{algorithm}

\section{Numerical experiments}
\label{sec:exp}
In this section, we present an experimental comparison of the BinNNMS to the $k$-modes clustering (as outlined in Section~\ref{sec:nnmsbin}).
Table~\ref{table:datasets} lists the details of the dataset obtained from the UCI Machine learning repository \cite{UCI}. The Zoo data set contains $n=101$ animals described with 16 categorical features: 15 of the variables are binary and one is numeric with 6 possible values. Each
animal is labelled 1 to 7 according to its class. Using disjunctive coding for the categorical
variable with 6 possible values, the data set consists of a $101 \times 21$ binary data matrix.
The Digits data concerns a dataset consisting of the handwritten numerals (``0''--``9'') extracted
from a collection of Dutch utility maps. There are 200 samples of each digit
so there is a total of $n=2000$ samples. As each sample is a $15\times16$ binary pixel image, the
dataset consisted of a $2000 \times 240$ binary data matrix. 
The Spect dataset describes the cardiac diagnoses from Single Proton Emission Computed Tomography (SPECT) images. Each patient is classified into two categories: normal and abnormal; there are $n=267$ samples which are described by 22 binary features.
%
The Car dataset contains examples with the structural information of the vehicle is removed. Each instance is classified into 4 classes.  This database is highly unbalanced since the distribution of the classes is $(70.02\%, 22.22\%, 3.99\%, 3.76\%)$. 
The Soybean data is about 19 classes, but only the first 15  have been ijustified as it appears that the last four classes are not well-defined. There are 35 categorical attributes, with both nominal and  ordinal features. 

\begin{table}[!htbp]
\centering
\setlength{\tabcolsep}{2pt}
\begin{tabular}{@{}|c|c|c|c|@{}}
\hline Dataset 		& size ($n$) & \#features ($d$) & \#classes ($M$) \\
\hline Zoo 		& 101  & 26	 & 7     \\ 
\hline Digits 	& 2000 & 240 & 10    \\ 
\hline Spect    & 267  & 22	 & 2     \\ 
\hline Soybean 	& 307  & 97	 & 18    \\ 
\hline Car 		& 1728 & 15	 & 4     \\ 
\hline 
\end{tabular} 
\caption{Overview of experimental datasets.}
\label{table:datasets}
\end{table}

\subsection{Comparison of the $k$-modes and the BinNNMS clustering}

To evaluate the clustering quality, we compare the known cluster labels in 
Table~\ref{table:datasets} to the estimated cluster labels from BinNNMS and $k$-modes. 
For comparability, the $k$-modes clustering is also based on the binary median center from Equation~\eqref{eqn:cost}.  Values of the Adjusted
Rand Index (ARAND) \cite{hubert1985} and the normalized mutual information
(NMI) \cite{strehl2002} close to one indicate highly matched cluster
labels, and values close to zero for the NMI/less than zero for the ARAND)
indicate mismatched cluster labels. 

Table \ref{tab:IndexesComparisonKMeansMS2} reports the results in terms of the NMI and ARAND after 10 runs of the BinNNMS and $k$-modes. Unlike BinNNMS, the $k$-modes clustering requires an a priori number of clusters $k$, then we set $k$ to be whichever value between the target number of classes from Table~\ref{table:datasets}, or  to be the number of clusters obtained from the BinNNMS clustering gives the highest clustering accuracy. 
The BinNNMS, apart from the Car dataset, outperforms the $k$-modes algorithm on Zoo, Digits, Spect, and Soybean datasets.
Upon further investigation for the Car dataset, recall that the distribution of the cluster labels is highly unbalanced which leads the BinNNMS giving a single class (i.e. no clustering). These unbalanced clusters also translate into 
low values of the NMI and ARAND for the $k$-modes clustering. 
%

\begin{table}[!htb]
\centering
\setlength{\tabcolsep}{3pt}
\begin{tabular}{@{}|l|cc|c|@{}}
\hline
&  \multicolumn{3}{c|}{NMI} \\ \hline
Dataset & $k$-modes & $k$ & BinNNMS  \\ \hline
Digits & $0.360 \pm 0.011$ & 40 & $\bf 0.880 \pm 0.000$ \\\hline
Zoo & $ 0.789 \pm 0.023$ & 8  & $\bf 0.945 \pm 0.000 $  \\\hline
Soybean & $ 0.556 \pm 0.000$ & 40 & $\bf 0.743 \pm 0.000 $ \\\hline
Spect &  $0.135 \pm 0.000$ & 47 & $\bf 0.145 \pm 0.000 $ \\\hline
Car  & $\bf{0.039 \pm 0.019}$ & $\bf{4}$ &  Single class \\\hline
& \multicolumn{3}{c|}{ARAND} \\ \hline
Dataset & $k$-modes & $k$ & BinNNMS \\ \hline
Digits & $0.166 \pm 0.021$ & 40 & $\bf 0.876 \pm 0.000 $ \\\hline
Zoo &  $ 0.675 \pm 0.032$ & 8 & $\bf 0.904 \pm 0.000$ \\\hline
Soybean & $ 0.178 \pm 0.000$ & 40 & $\bf 0.331 \pm 0.000 $ \\\hline
Spect & $\bf{-0.009 \pm 0.055}$ & $\bf{2}$ & $ -0.019 \pm 0.000 $ \\\hline
Car  & $\bf{0.016 \pm 0.039}$ & $\bf{4}$ &  Single class\\\hline
\end{tabular}
\caption{Comparison of clustering quality indices (NMI and ARAND) for $k$-modes and BinNNMS. The bold value indicates the most accurate clustering for the dataset.}
\label{tab:IndexesComparisonKMeansMS2}
\end{table}

\subsection{Comparison of the tuning parameters for the BinNNMS clustering}
\begin{figure*}[!htb]
\centering
\begin{tabular}{@{}c@{}c@{}}
\multicolumn{2}{c}{Digits} \\ 
\includegraphics[height=5cm,width=0.8\columnwidth]{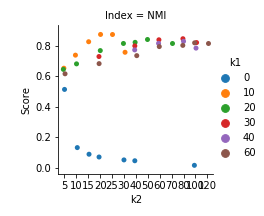} &
\includegraphics[height=5cm,width=0.8\columnwidth]{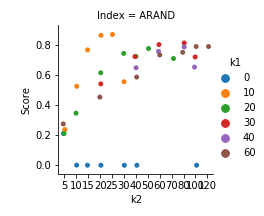} \\
\multicolumn{2}{c}{Zoo} \\ 
\includegraphics[height=5cm,width=0.8\columnwidth]{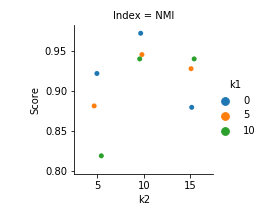} &
\includegraphics[height=5cm,width=0.8\columnwidth]{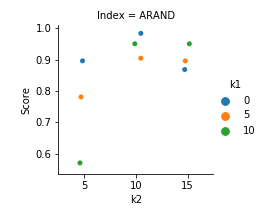} \\
\multicolumn{2}{c}{Soybean} \\
\includegraphics[height=5cm,width=0.8\columnwidth]{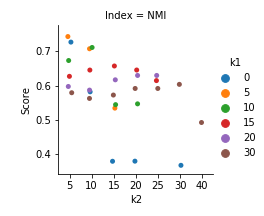} &
\includegraphics[width=0.8\columnwidth]{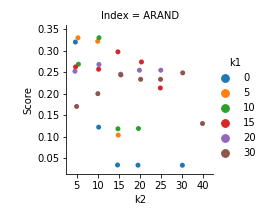} \\
\multicolumn{2}{c}{Spect} \\
\includegraphics[height=5cm,width=0.8\columnwidth]{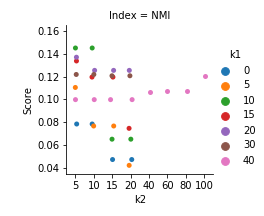} &
\includegraphics[height=5cm,width=0.8\columnwidth]{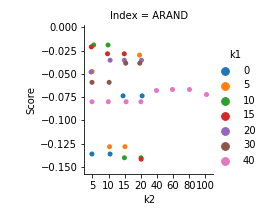}
\end{tabular}
\caption{Evolution of the cluster quality indices (NMI and ARAND) as functions of the $k_1$ and $k_2$ tuning parameters for the BinNNMS
for the Digits, Zoo, Soybean and Spect datasets.}
\label{fig:ms-index-evolution}
\end{figure*}

Figure~\ref{fig:ms-index-evolution} presents the evolution of the NMI and ARAND scores as a function of the tuning parameters $k_1$ (in binary gradient ascent BGA task) and $k_2$ (in the cluster labeling task) for 
the Digits, Soybean and Spect datasets. The blue dots ($k_1=0$) correspond to the application of the cluster labeling task  without the gradient ascent. These cases tend to have poor cluster quality values compared to when
$k_1$ is non-zero. Otherwise, that various values of $k_1$ and $k_2$ give the highest
cluster label accuracy indicate that the optimal combinations of these tuning parameters remains an open and challenging task. 

%

%
%
%
%
%
%
%

\subsection{Comparison of the quantization errors for the BinNNMS}
An important and widely used measure of resolution, the quantization error, is computed based on Hamming distances
between the data points and the cluster prototypes: 
\begin{equation}
Error = \frac{1}{n}\sum_{m=1}^M {\sum_{\bx_j \in \mathcal C_m}{{\cal H}(\bx_j, \bw_m)}}
\end{equation}
where $\{\mathcal C_1, \dots, \mathcal C_M\}$ is the set of $M$ clusters, $\bx$ is a point assigned to cluster $\mathcal C_m$, 
and $\bw_m$ is the prototype.median center of cluster $\mathcal C_m$.

\begin{figure*}[!htp]
\centering
\setlength{\tabcolsep}{0pt}
\begin{tabular}{@{}cc@{}}
\multicolumn{2}{c}{Digits} \\ 
\includegraphics[width=5.5cm,height=0.2\textheight]{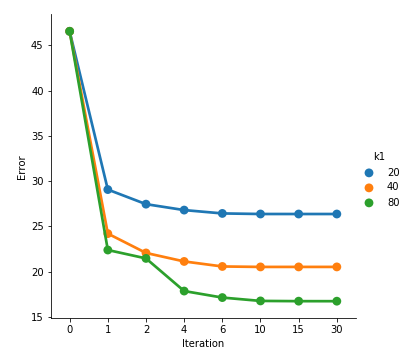} &
\includegraphics[width=6.4cm,height=0.2\textheight]{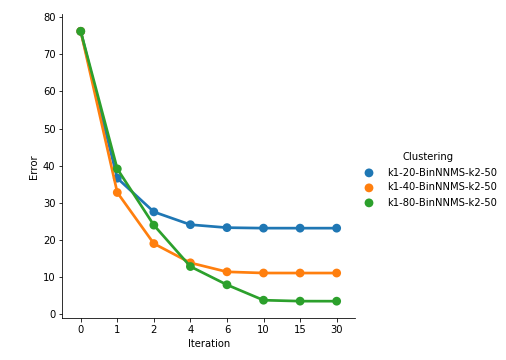} \\
\multicolumn{2}{c}{Zoo} \\ 
\includegraphics[width=5.5cm,height=0.2\textheight]{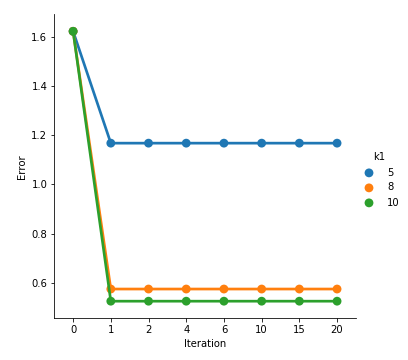} &
\includegraphics[width=6.4cm,height=0.2\textheight]{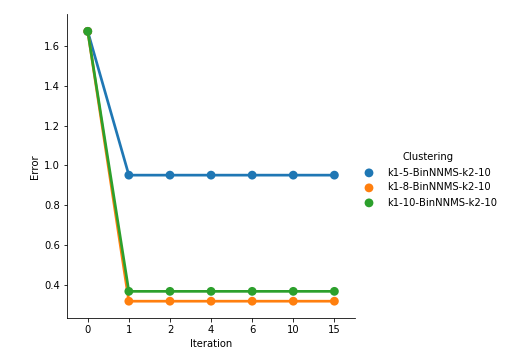} \\
\multicolumn{2}{c}{Soybean} \\ 
\includegraphics[width=5.5cm,height=0.2\textheight]{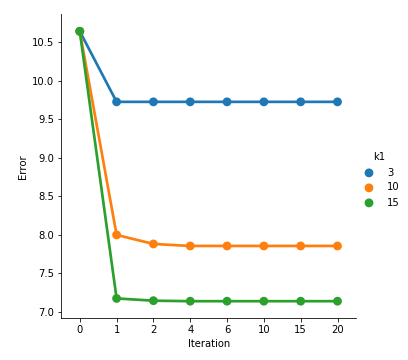} &
\includegraphics[width=6.4cm,height=0.2\textheight]{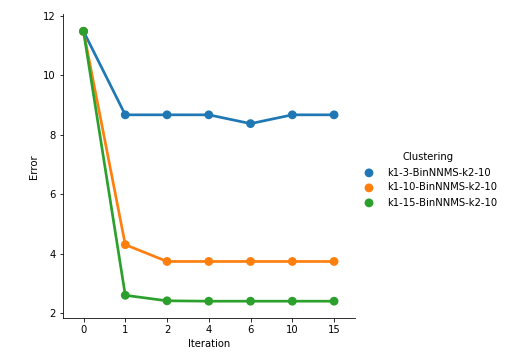} \\
\multicolumn{2}{c}{Spect} \\ 
\includegraphics[width=5.5cm,height=0.2\textheight]{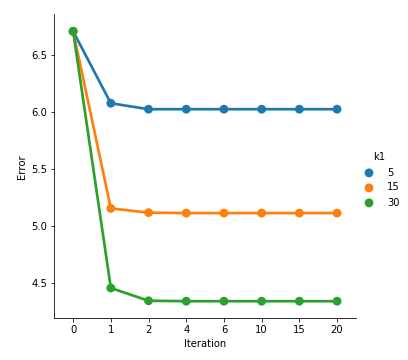}
&
\includegraphics[width=6.4cm,height=0.2\textheight]{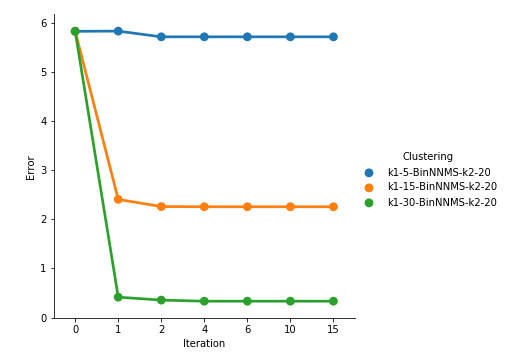}
\end{tabular}
\caption{Evolution of quantization errors as a function of the $k_1$ and $k_2$ tuning parameters in BinNNMS
for the Digits, Zoo, Soybean and Spect datasets. Left. Quantization errors between the data points and the target prototypes. 
Right. Quantization errors between the data points and the intermediate median centers in the BGA task and the cluster prototypes.}
\label{fig:error-evolution-on-GT-over-iterations}
\end{figure*}

The right hand column in Figure~\ref{fig:error-evolution-on-GT-over-iterations} 
shows the evolution of the quantization errors for the BinNNMS with different values of $k_1$ with respect to
the target cluster prototypes. 
As the quantization errors decrease this implies that the data points converge toward their cluster prototypes, and that the decreasing intra-cluster distance further facilitates the clustering process. Thus at the end of the training phase, the data points converge towards to their local mode.  
In comparison with the ARAND scores in Table~\ref{tab:IndexesComparisonKMeansMS2}, the magnitude of the 
decrease in the quantization errors is inversely proportional to the cluster quality indices. That is, the 
largest decrease for the Digits dataset implies that BinNNMS clustering achieves here the highest ARAND score.

If we run the labeling phase during the BGA phase for a fixed $k_1$ then we compute the intermediate prototypes $\bw_m$ of the  clusters  $\mathcal C_m$ during  the binary gradient ascent BGA task. 
Since BinNNMS provides clusters as the basins of attraction to the local median created by the binary gradient ascent paths, 
the left column of Figure \ref{fig:error-evolution-on-GT-over-iterations} shows the quantization error  
with respect to the intermediate median centers/prototypes. In this case we compute  at each iteration 7 modes for Zoo dataset,  
10 modes for the Digits, 18 modes for Soybean and 2 modes for Spect datasets using ground truth. 
These quantization errors decrease to an asymptote for all datasets as the iteration number increases.

\subsection{Visual comparison of $k$-modes and BinNNMS on the Digit dataset}
\begin{figure*}[!h]
\centering
\begin{tabular}{cc} 
$k$-modes & BinNNMS \\
\includegraphics[width=0.7\columnwidth]{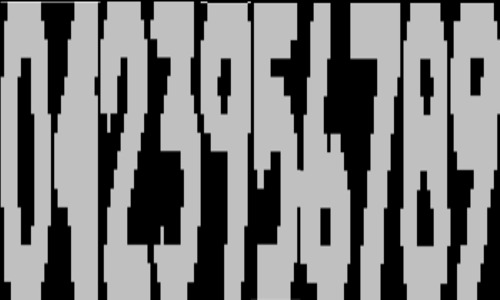} &
\includegraphics[width=0.7\columnwidth]{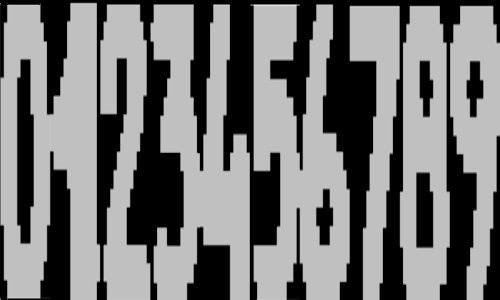}
\end{tabular}
\caption{Comparison of the $k$-modes and BinNNMS clustered images for the Digits dataset.}
\label{fig:kmodesVsBNNMS}
\end{figure*}

%
%
Figure \ref{fig:kmodesVsBNNMS} show the cluster prototypes provided by $k$-modes and BinNNMS, displayed as $15\times16$ binary pixel images. For the $k$-modes image, the cluster prototype for the ``4'' digit has been incorrectly associated
with the ``9'' cluster. On the other hand, the BinNNMS image correctly identifies all ten digits from ``0'' to ``9''.  

\section{Conclusion}
In this paper, we have proposed a new and efficient modal clustering method for binary data. We introduced a  mathematical analysis of the nearest neighbor estimators for binary data. This was then combined with the  
Aitchison and Aitken kernel in order to generalize the traditional mean shift clustering to the 
median shift clustering for binary data (BinNNMS).
Experimental evaluation for a number of experimental datasets demonstrated that the BinNNMS outperformed 
the $k$-modes clustering in terms of visual criteria, as well as quantitative clustering quality criteria such as the adjusted Rand index, the normalized mutual information and the quantization error. In the future we envisage to make our algorithm as automatic as possible by optimizing the choice of the tuning parameters, and to implement a scalable version for Big Data by using approximate nearest neighbor searches.

\bibliographystyle{unsrt}
\bibliography{biblio}

\begin{thebibliography}{10}

\bibitem{Huang97clusteringlarge}
Z.~Huang.
\newblock Clustering large data sets with mixed numeric and categorical values.
\newblock In {\em The First Pacific-Asia Conference on Knowledge Discovery and
  Data Mining}, pages 21--34, 1997.

\bibitem{macqueen1967}
J.~MacQueen.
\newblock Some methods for classification and analysis of multivariate
  observations.
\newblock In {\em Proceedings of the Fifth Berkeley Symposium on Mathematical
  Statistics and Probability, Volume 1: Statistics}, pages 281--297, Berkeley,
  USA, 1967. University of California Press.

\bibitem{Li2006}
T.~Li.
\newblock A unified view on clustering binary data.
\newblock {\em Mach. Learn.}, 62:199--215, 2006.

\bibitem{LebbahBT00}
M.~Lebbah, F.~Badran, and S.~Thiria.
\newblock Topological map for binary data.
\newblock In {\em {ESANN} 2000, 8th European Symposium on Artificial Neural
  Networks, Bruges, Belgium, April 26-28, 2000, Proceedings}, pages 267--272,
  2000.

\bibitem{fukunaga1973}
K.~Fukunaga and L.~Hostetler.
\newblock Optimization of $k$-nearest-neighbor density estimates.
\newblock {\em {IEEE} Trans. Inform. Theory}, 19:320--326, 1973.

\bibitem{fukunaga1975}
K.~Fukunaga and L.~Hostetler.
\newblock The estimation of the gradient of a density function, with
  applications in pattern recognition.
\newblock {\em {IEEE} T. Inform. Theory}, 21:32--40, 1975.

\bibitem{Hamming50}
R.~W. Hamming.
\newblock Error detecting and error correcting codes.
\newblock {\em Bell Syst. Tech. J.}, 29:147--160, 1950.

\bibitem{loftsgaarden1965}
D.~O. Loftsgaarden and C.~P. Quesenberry.
\newblock A nonparametric estimate of a multivariate density function.
\newblock {\em Ann. Math. Statist.}, 36:1049--1051, 06 1965.

\bibitem{duong2016prl}
T.~Duong, G.~Beck, H.~Azzag, and M.~Lebbah.
\newblock Nearest neighbour estimators of density derivatives, with application
  to mean shift clustering.
\newblock {\em Pattern Recogn. Lett.}, 80:224--230, 2016.

\bibitem{Leich98}
F.~Leisch, A.~Weingessel, and E.~Dimitriadou.
\newblock Competitive learning for binary valued data.
\newblock In L.~Niklasson, M.~Bod{\'e}n, and T.~Ziemke, editors, {\em {ICANN}
  98}, pages 779--784, London, 1998. Springer.

\bibitem{Diday1976}
E.~Diday and J.~C. Simon.
\newblock {\em Clustering Analysis}, pages 47--94.
\newblock Springer, Berlin, 1976.

\bibitem{aitchinson1976}
J.~Aitchison and C.~G.~G. Aitken.
\newblock Multivariate binary discrimination by the kernel method.
\newblock {\em Biometrika}, 63:413--420, 1976.

\bibitem{UCI}
D.~Dheeru and E.~Karra~Taniskidou.
\newblock {UCI} machine learning repository, 2017.

\bibitem{hubert1985}
L.~Hubert and P.~Arabie.
\newblock Comparing partitions.
\newblock {\em J. Classif.}, 2:193--218, 1985.

\bibitem{strehl2002}
A.~Strehl and J.~Ghosh.
\newblock Cluster ensembles -- a knowledge reuse framework for combining
  multiple partitions.
\newblock {\em J. Mach. Learn. Res.}, 3:583--617, 2002.

\end{thebibliography}

\end{document}